\pdfoutput=1
\documentclass[11pt]{article}
\usepackage[usenames,dvipsnames]{xcolor}
\usepackage{acl}

\usepackage{times}
\usepackage{latexsym}

\usepackage[T1]{fontenc}
\usepackage[utf8]{inputenc}

\usepackage{microtype}

\usepackage{inconsolata}

\usepackage[utf8]{inputenc} % allow utf-8 input
\usepackage[T1]{fontenc}    % use 8-bit T1 fonts
\usepackage{color,xcolor}
\usepackage{hyperref}       % hyperlinks
\usepackage{url}            % simple URL typesetting
\usepackage{booktabs}       % professional-quality tables
\usepackage{amsfonts}       % blackboard math symbols
\usepackage{nicefrac}       % compact symbols for 1/2, etc.
\usepackage{microtype}      % microtypography
\usepackage{graphicx}       %% SS added for framework fig import
\usepackage{multirow}
\usepackage{graphicx}
\usepackage{tabularx}
\usepackage{makecell}
\usepackage{amsmath}
\usepackage{amsthm}

\usepackage{graphicx} %% SS added for framework fig import
\usepackage{multirow}
\usepackage{graphicx}
\usepackage{makecell}       % 单元格内换行
\usepackage{arydshln}

\usepackage{amssymb}
\usepackage{bbding}
\usepackage{pifont}
\usepackage{wasysym}
\usepackage{utfsym}
\usepackage{fontawesome}
\usepackage{wrapfig}
\definecolor{darkpink}{rgb}{0.91, 0.33, 0.5}

\title{Controllable Text Generation  with Residual Memory Transformer}

\author{Hanqing Zhang$^{1*}$, Si Sun$^{2*}$, Haiming Wu$^{1}$, Dawei Song $^{1\dagger}$ \\
$^{1}$School of Computer Science \& Technology, Beijing Institute of Technology \\
$^{2}$Department of Electronic Engineering, Tsinghua University \\
\texttt{zhanghanqing@bit.edu.cn} \text{ } \texttt{s-sun17@mails.tsinghua.edu.cn} \\
\texttt{haimming@bit.edu.cn} \text{ } \texttt{dwsong@bit.edu.cn} \\
$^{*}$Equal contributions $^{\dagger}$Corresponding author
}

\begin{document}
\maketitle

\begin{abstract}

Large-scale Causal Language Models (CLMs), e.g., GPT3 and ChatGPT, have brought great success in text generation. However, it is still an open challenge to control the generation process of CLM while balancing flexibility, control granularity, and generation efficiency. In this paper, we provide a new alternative for controllable text generation (CTG), by designing a non-intrusive, lightweight control plugin to accompany the generation of CLM at arbitrary time steps. The proposed control plugin, namely Residual Memory Transformer (RMT), has an encoder-decoder setup, which can accept any types of control conditions and cooperate with CLM through a residual learning paradigm, to achieve a more flexible, general, and efficient CTG. Extensive experiments are carried out on various control tasks, in the form of both automatic and human evaluations. The results show the superiority of RMT over a range of state-of-the-art approaches, proving the effectiveness and versatility of our approach. \href{https://github.com/littlehacker26/Residual_Memory_Transformer}{Click here for source code.}

\end{abstract}

\section{Introduction}
% Controllable text generation (CTG) focuses on generating text while adhering to specific constraints~\cite{NEURIPS2021_d0f5edad, ctg_survey}. These constraints can range from high-level semantic elements, such as emotions, topics, and toxicity avoidance, to finer-grained content, e.g., including specific concepts or key elements in the generated text. This capability is critical in real-world applications, such as personalized chatbots~\cite{zheng2020pre,zhong-etal-2022-less}, story generation~\cite{10.1145/3340531.3411937}, and the development of safe AI applications~\cite{gedi,liu-etal-2021-dexperts}.

Controllable text generation (CTG) focuses on generating text while adhering to specific constraints~\cite{NEURIPS2021_d0f5edad, ctg_survey}. These constraints can range from high-level semantic elements, such as emotions, topics, and toxicity avoidance, to finer-grained content, e.g., including specific concepts or key elements in the generated text. With the development of generative AI based on large language models, social media will be flooded with AI-generated content. Therefore, CTG will be critical in the real-world Web applications to establish the safer, more reliable and practical~\cite{gedi,liu-etal-2021-dexperts, ctg_survey}  AI-driven social systems.

The state-of-the-art CTG methods are based on Large Language Models (LLMs) that build upon the Transformer structure~\cite{transformer} and have gained significant attention due to their remarkable ability to understand and generate text. Recently, large-scale causal Language Models (CLMs), i.e., decoder-only language models, show particular advantages in zero/few-shot scenarios~\cite{gpt3}, resulting in a series of successors such as ChatGPT and GPT4. This has been seen as a milestone towards the realization of Artificial Generative Intelligence. Despite the success of these models, CLMs are still facing certain challenges, especially in CTG.

Considering the significant scale of CLMs and the substantial cost of training such models, current mainstream CLM-based CTG methods fall into two categories, i.e., prompt-based and post-processing approaches. The prompt-based methods~\cite{zhang-song-2022-discup, tailor_2022, contrast_prefix, lu2022quark,pmlr-v202-zhou23g} concatenate the control-prompt with the input head of the generative model to instruct more controllable text generation. As previous studies have revealed~\cite {inverse_prompt, carlsson-etal-2022-fine}, the control effectiveness tends to deteriorate with the increasing distance from the prompt. Additionally, inserting a control-prompt into a well-trained model may harm the model's original generative stream, thus losing the flexibility of control~\cite{carlsson-etal-2022-fine}. On the other hand, most post-processing methods leverage an auxiliary module to adjust the probability of naturally producing a token by the generative model at the decoding phase~\cite{gedi,liu-etal-2021-dexperts,yang-klein-2021-fudge, lu-etal-2022-neurologic}, hindering the model's capacity of content planning and thus limiting the fine-gained control. Furthermore, more recent decode-time methods~\cite{li2022diffusion, mireshghallah-etal-2022-mix,qin2022cold} improve the control granularity through iterative sampling or editing, but at the expense of generation efficiency. Therefore, the flexibility, control granularity and generation efficiency need to be better balanced, which demands a more versatile CLM-based CTG framework\footnote{We conclude current CTG methods in Appendix~\ref{CTG_Approaches_Comparison}.}.

In this paper, we propose a novel CTG plugin named Residual Memory Transformer (RMT), which borrows the paradigm of residual learning~\cite{resnet, zhang2020side} and only makes late fusion with a frozen CLM to noninvasively steer the generation process. Unlike prompt-based approaches, this paradigm does not disturb the original generative stream of the base CLM model, allowing for a better flexibility of CTG (i.e., control with a  plug-and-play manner). In addition, the RMT architecture consists of an encoder-decoder structure, where the encoder handles different types of control information and influences the generation process, so as to achieve fine-grained control. Meanwhile, RMT utilizes cross-attention to uniformly apply control conditions to each generated token, avoiding the negative effect varying with context length. In particular, different from the vanilla decoder of Transformer, an additional causal attention is introduced to extract the prior knowledge from the generative stream of CLM, allowing RMT not to deviate too far away from the original generative model while its implied high-level semantics is leveraged. The reuse of the base-CLM's output allows RMT to achieve effective control with a tiny network, resulting in an improved generation efficiency.

The training of RMT includes two stages: pre-training and fine-tuning. The pre-training of RMT aims to reconstruct noisy text into a complete sentence, facilitating RMT's understanding of the semantics of various control conditions, while aligning with the generative process of CLM. During the fine-tuning stage (i.e., residual learning), the logit of the RMT is directly added to that of the fixed CLM, and the goal of training is to learn the parameters of RMT such that the joint model distribution gets close to the desired text distribution. Since the gradient does not need to be backpropagated to base-CLM, the training of RMT is very efficient. For instance, in our experiments based on GPT2-large, the entire pre-training stage of RMT with 4M  sample could be completed in approximately 30 hours, using a single NVIDIA A6000 GPU.

We conduct extensive experiments to explore the superiority of our approach in three aspects. \textbf{(1) Flexibility:} Theoretically, the proposed RMT has the capability to intervene in the generation process at any step. Experimentally, it maintains the same level of control effectiveness in both with-context and without-context settings, showing that RMT enables long-distance control throughout the generation process. \textbf{(2) Control Granularity:} We test our approach on a range of CTG tasks of different granularity levels (i.e., fine-gained control tasks including word inclusion and sentence length control; and attribute control based on sentiment). RMT achieves a control effectiveness comparable to the state-of-the-art approaches, while guaranteeing the text quality. \textbf{(3) Efficiency:} The results show that three-layer blocks of RMT is enough to achieve a competitive CTG effectiveness, and the time-cost of text generation is almost the same to the original CLM.

% Our contributions are summarized as followings: 

% (1) We propose an novel and general controllable text generation framework, which leverages an smaller scale encoder-encoder architecture to steer the CLM by a plug-and-play manner, enabling more flexible and efficient controllability in text generation, without modifying and tuning the CLM itself. (2) We propose a novel control model, namely Residual Memory Transformer, which  synthesizes multi-source information such as control and contextual streams to effectively control the generation of CLMs.
% (3) We conduct extensive experiments on different-level CTG (i.e., Word Inclusion, Sentence Length Control, and Attribute control), and the results prove the effectiveness and superiority of our proposed method.

\section{Preliminary}
\label{gen_inst}

\subsection{Attention in Transformer}

\textbf{Self-Attention} mechanism is a critical component in Transformer~\cite{transformer}, which is used to effectively capture long-range dependencies between words in the input sequence. Specifically, it is defined as a mapping between a query and a collection of key-value pairs. The values are weighted according to a compatibility function between the query and each corresponding key, and then summed up,  eventually obtaining the output vector. The self-attention mechanism can be formatted as follows:

\begin{equation}
\operatorname{Att}(Q, K, V)=\operatorname{softmax}\left(\frac{Q K^T}{\sqrt{d_k}}\right) V, \label{eq.self-attn}
\end{equation}

where $Q,K,V$ represent the transformed representations of the sequence using the corresponding learnable matrix, and $\sqrt{d_k}$ is the scaling factor. In the Transformer model, the key, query and value vectors are always derived from a shared sequence.

\textbf{Causal Attention} is a particular branch of self-attention, also called masked self-attention, which is usually applied in the decoder module of Transformer. Different from normal self-attention, the queries in causal attention are only confined to their preceding key-value pairs' positions and their current position to maintain the auto-regressive property. Usually, it can be implemented by a mechanism that masks the invalid positions and sets them to negative infinite:

\begin{gather}
% \begin{align}
\operatorname{Att}(Q, K, V)=\operatorname{softmax}\left(\frac{QA K^T}{\sqrt{d_k}}\right) V,
\label{eq.causal-attn}  \\
\begin{aligned}
{{A}_{ij}}=\left\{ \begin{aligned}
& 1\text{       }\text{if}\text{ }i\ge j, \\ 
 & -\infty \text{  }\text{else}. \\ 
\end{aligned} \right.
\end{aligned}
% \end{align}
\end{gather}

\textbf{Cross Attention} is another branch of self-attention in the decoder part of Transformer, which aims to capture the interaction between the encoder and decoder. Specifically, The key and value parts are obtained from the previous step outputs of the encoder, and the query is from the decoder, enabling the encoder to attend to each position in the decoder module at the same time.

\subsection{Causal Language Model}

\textbf{Causal Language Model (CLM)} eliminates the need for a separate encoder component and typically leverages a decoder-only Transformer~\cite{transformer}, in which only the causal attention is used to perform step-wise density estimation, i.e., predicting the next token.  Suppose a CLM  is parameterized with $\theta$. Given a partial sequence $x_{<t}$, it assigns a probability  $P_{\theta}(x_{t}| x_{<t})$ over a vocabulary $\mathcal{V}$ for next-token $x_t$ generation. When generating a sequence of text $X_{n}=\{x_1, x_2, \dots, x_n \}$,  it can be formulated by the chains rule as below:

\begin{equation}
\label{forward}
P_{\theta}\left(X_{n}\right)=\prod_{t=1}^{n} P_{\theta}\left(x_{t} \mid x_{<t}\right).
\end{equation}

The entire generation process is carried out iteratively. First, a token is sampled from the distribution $P_{\theta}\left(x_{t} \mid x_{<t}\right)$. Then the selected token is concatenated with the input for the next step of generation.

% \section{Problem Definition}
%  % Given a trained causal Language Model (CLM), e.g., GPT family models, we hope to establish a noninvasive control mechanism, to steer it generate texts satisfying the certain constraint (e.g., sentiment, topic, toxicity, keywords, etc.). 
%  Unlike traditional conditional text generation (i.e., using a prompt as a control condition to instruct CLM), the control paradigm defined in our paper is similar to the post-process manner, where original CLMs condition on nothing. And an auxiliary control model could be plugged/unplugged at an arbitrary generation step. If the control module is activated, it can steer the CLM to generate text toward desired conditions. When deactivated, the CLM will generate text as usual, as per its training. It can be formally described as:

%  \begin{equation}
% P_{c}\left(x_{t} \mid x_{<t}, c\right) \propto P_{L M}\left(x_{t} \mid x_{<t}\right) E_{\theta}\left(x_{<t},C \right),
% \label{bays_rules}
% \end{equation}
% Where $P_{c}\left(x_{t} \mid x_{<t}, C\right)$ is regarded as the final probability for text generation, $P_{L M}$ is the trained CLM which is fixed during the control model training, $E_{\theta}$ is the control model and $C$ represents the control condition varying according to different tasks. The above-defined control paradigm is considered more challenging than prompt-based approaches; however, since the control module is separated from the CLM, it is more flexible and practical.

%% **************************************
%% Framework Fig
\begin{figure*}[t]
\centering
\includegraphics[width=\textwidth]{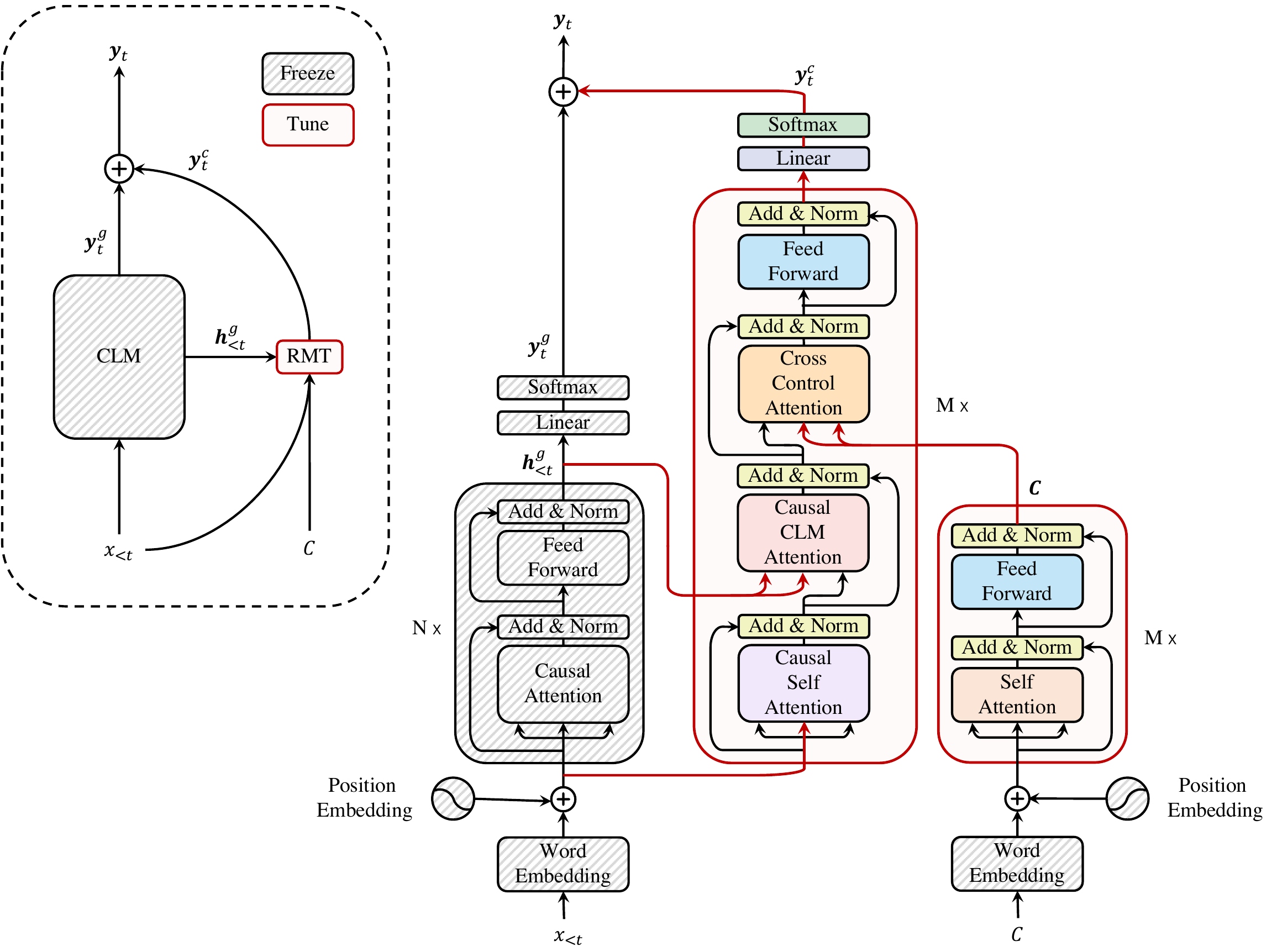}
\caption{\label{fig:framework} Illustration of controllable text generation with Residual Memory Transformer (RMT). In the top left, we present a miniaturization of our method framework, where the gray sector represents the frozen CLM, and the red box symbolizes our plug-in control module, i.e., RMT. The right part illustrates the embodiment of each detailed component.}
\vspace{-0.2cm}
\end{figure*}
\vspace{-0.02cm}
%% **************************************

%% **************************************
\section{Methodology}
\label{headings}

This section introduces the proposed Residual Memory Transformer (RMT) to control the text generation with a causal language model (CLM). We first provide an overview of the RMT-enhanced controllable text generation (CTG) framework, followed by a detailed description of RMT, and finally, the dedicated pre-training and fine-tuning methods for the proposed framework.

\subsection{CTG Framework with RMT}
% This CTG paradigm not only ensures CLM's high-quality generation but also meets controllable conditions. 
As the dashed diagram in Figure~\ref{fig:framework} illustrates, RMT operates in a non-intrusive control mode as a CLM plug-in, where the control signal passes through RMT independently and affects the output distribution of a frozen CLM through residual learning ($y_t = y^{g}_{t} + y^{c}_{t}$) without interfering with the CLM's free-generation passageway. Compared to intrusive controllable paradigms, e.g., prompt-based approaches, RMT allows for more flexibility in switching between freestyle generation and controllable generation modes. The proposed non-intrusive control paradigm is considered more challenging. Specifically, RMT is decoupled from the generation process of base-CLM  has the potential for dynamic control intervention. Meanwhile, it also avoid tuning the base-CLM into a task-specific model, ensuring the universality of the base-CLM. Hence, it is more flexible and promising. More explanation could be see in Appendix~\ref{CTG_Approaches_Comparison}.

Formally, given a partial generated text $x_{<t}$ and a sequence of control instruction $C$, the proposed framework aims to generate eligible text $X_{n} = \{x_1, ..., x_n\}$ that meets the control conditions:
\begin{equation}
P_{\Theta}\left(X_{n}\right)=\prod_{t=1}^{n} P_{\Theta}\left(x_{t} \mid x_{<t} ; C \right),
\end{equation}
where $\Theta = \{\tilde{\theta}; \phi \}$ represents the CTG model's parameters, which comprise both the frozen parameters inherited from the CLM ($\tilde{\theta}$) and the tunable parameters derived from the RMT ($\phi$). The entire generation process of our approach consists of the following four steps:

\textbf{CLM's Raw Generation.} At the generation step $t$, CLM first maps the generated text $x_{<t} = \{x_1, ..., x_{t-1}\}$ to hidden states $\mathbf{h}^{g}_{<t} = \{\mathbf{h}^{g}_1, ..., \mathbf{h}^{g}_{t-1}\}$. Afterward, a linear layer and softmax normalization are used to project the hidden state of the last token $\mathbf{h}^{g}_{t-1}$ to the probability distribution $\mathbf{y}^{g}_t$ over the vocabulary:
\begin{equation}
\mathbf{y}^{g}_t = P_{\tilde{\theta}}\left(x_{t}|x_{<t}\right) = \text{softmax}(\text{Linear}(\mathbf{h}^{g}_{t-1})), \label{eq.clm-gen}
\end{equation}

where the CLM naturally generates the next token $x_t$ given the context $x_{<t}$ as per its training.

% where the higher the probability of a token over the vocabulary, the more likely it is to be selected by the CLM to generate the next token $x_t$ given the context $x_{<t}$.

\textbf{RMT's Control Encoding.} Next, the RMT's encoder is responsible for encoding the control instruction $C = \{c_1, ..., c_m\}$ into the control memory $\mathbf{C} = \{\mathbf{c_1}, ..., \mathbf{c_m}\}$, which is used to guide the controllable text generation in the RMT's decoder:
\begin{equation}
\mathbf{C} = \text{RMT-Enc}(C; \phi_{\text{enc}}). \label{eq.rmt-enc}
\end{equation}

\textbf{RMT's Control Decoding.} After encoding the control signal, the RMT's decoder maps the generated text $x_{<t}$ to new hidden states $\mathbf{h}^{c}_{<t} = \{\mathbf{h}^{c}_1, ..., \mathbf{h}^{c}_{t-1}\}$ by considering both the control memory $\mathbf{C}$ and CLM's vanilla hidden states $\mathbf{h}^{g}_{<t}$, and synthetically predicting next token's probability distribution $\mathbf{y}^{c}_t$ over the vocabulary:
\begin{equation}
\begin{aligned}
\mathbf{h}^{c}_{<t} = \{\mathbf{h}^{c}_1..., \mathbf{h}^{c}_{t-1}\} = \text{RMT-Dec}(x_{<t}, \mathbf{h}^{g}_{<t}, \mathbf{C}; \phi_{\text{dec}}), 
\\
\mathbf{y}^{c}_t = P_{\phi}\left(x_{t}|x_{<t};C\right) = \text{softmax}(\text{Linear}(\mathbf{h}^{c}_{t-1})). \label{eq.rmt-gen}
\end{aligned}
\end{equation}

% To incorporate both CLM's prior knowledge and control information,

\textbf{Residual Learning to Generate.}  We employ residual learning to fuse the output distributions from the CLM (Eq.~\ref{eq.clm-gen}) and the RMT (Eq.~\ref{eq.rmt-gen}), allowing the framework to obtain the joint predictions for the next token and achieve a noninvasive CTG:
\begin{equation}
\mathbf{y}_t = P_{\Theta}\left(x_t|x_{<t};C\right) = \mathbf{y}^{g}_t + \mathbf{y}^{c}_t.
\end{equation}

\subsection{Residual Memory Transformer (RMT)}

% \sunsi{1. get more connection to beforhand func. 2. give explain for three attention 3. revise figure 1.}

The detailed structure of RMT is shown in Figure~\ref{fig:framework}. Specifically, RMT adopts an encoder-decoder architecture and reuses the CLM's suite of word and position embeddings.

\textbf{RMT Encoder.} The RMT encoder is responsible for encoding the control description $C$ into the control memory $\mathbf{C}$ (Eq.~\ref{eq.rmt-enc}). The encoder is composed of a stack of $M$ identical blocks. Each block comprises a self-attention layer (Eq.~\ref{eq.self-attn}) and a fully connected feed-forward layer. Additionally, it incorporates residual connections around above each layer, and is followed by a layer normalization.

\textbf{RMT Decoder.} The RMT decoder aims to predict the probability distribution of the next token $\mathbf{y}_{t}^{c}$ (Eq.~\ref{eq.rmt-gen}) in the control mode. The decoder is also composed of a stack of $M$ identical blocks. Each block contains three carefully-designed attention layers and a fully connected feed-forward network. Similar to the encoder, residual connections and layer normalization are also applied. The three attention layers are directed at the already generated text $x_{<t}$, the CLM's output hidden states $\mathbf{h}_{<t}^{g}$, and the control memory $\mathbf{C}$, respectively:
\begin{itemize}
  \item \textbf{\textit{Causal Self Attention}}. The first attention layer utilizes a causal attention operation (Eq.~\ref{eq.causal-attn}), whereby $Q$, $K$, and $V$ are all firstly mapped from the original generated text $x_{<t}$ itself. This attention mechanism facilitates the identification and capturing contextual features of generated sequence from scratch.

  \item \textbf{\textit{Causal CLM Attention}}. The second attention layer also employs causal attention (Eq.~\ref{eq.causal-attn}), but with a key difference: $Q$ is sourced from the previous causal self-attention layer's output, while $K$ and $V$ are obtained from the CLM's last hidden states $\mathbf{h}_{<t}^{g}$. This design establishes an inner residual connection with CLM, enabling RMT to consider the high-level contextual features and maximally interact with the generative stream of CLM.

  \item \textbf{\textit{Cross Control Attention}}. The third attention layer is a cross-attention for the control memory from the RMT encoder. Specifically, $Q$ is sourced from the previous causal CLM attention layer, while $K$ and $V$ are derived from the control memory $\textbf{C}$. This cross-attention layer bridges the RMT's encoder and decoder, introducing the control signals to the generation.

\end{itemize}

\subsection{Model Training}
% RMT is an encoder-decoder architecture featuring a bidirectional encoder and an auto-regressive decoder. 
\textbf{Pre-training.} To enable the RMT's  semantic understanding capability and allow it align with the CLM's generation process, we utilize the denoising auto-encoder pre-training method, whereby the encoder processes the corrupted text $\hat{X}$ and the decoder reconstructs the original text $X$.
% \begin{equation}
% \begin{aligned}
% \hat{\mathbf{X}} &= \text{RMT-Enc}(\hat{X}; \phi_{\text{enc}}),
% \\
% X &= \text{RMT-Dec}(\hat{X}, \hat{\mathbf{X}}; \phi_{\text{dec}}). \label{eq.rmt-pretrin}
% \end{aligned}
% \end{equation}
% \begin{equation}
% \begin{aligned}
% X \rightarrow \text{} &\hat{X},
% \\
% \text{RMT-Enc}(\hat{X}; \phi_{\text{enc}}) = \text{} &\hat{\mathbf{X}},
% \\
% \text{RMT-Dec}(\hat{X}, \hat{\mathbf{X}}; \phi_{\text{dec}}) = \text{} &X. \label{eq.rmt-pretrin}
% \end{aligned}
% \end{equation}

Specifically, we corrupt the pretraining text ($X \rightarrow \hat{X}$) in four ways referring to~\cite{lewis-etal-2020-bart}: \textit{(1) Token Masking:} randomly masking tokens with special tokens. \textit{(2) Token Deletion:} randomly deleting tokens in the text. \textit{(3) Span Replacing:} randomly replacing the text spans of different lengths with special tokens. Different from token masking, each span is replaced with a single special token. \textit{(4) Text Rotation:} randomly selecting a token and rotating the text around it, i.e., the text begins with that token and ends with its previous token. More details are supplied in Appendix~\ref{pretraining_details}.

% \begin{itemize}
%   \item \textbf{\textit{Token Masking.}} Randomly mask tokens with special tokens.
%   \item \textbf{\textit{Token Deletion.}} Randomly delete tokens and mark their positions.
%   \item \textbf{\textit{Span Replacing.}} Randomly replace text spans of different lengths with special tokens. different from token masking, each span is replaced with a single special token.
%   \item \textbf{\textit{Text Rotation.}} Randomly select a token and rotate the text around it, i.e., the text begins with that token and ends with its previous token.
% \end{itemize}

\textbf{Fine-tuning.} Once the pre-training is complete, RMT can be fine-tuned to perform various controllable text generation tasks. This allows the CLM to generate sequences that meet specific control requirements in an auto-regressive manner. The specific objective can vary according to different tasks: we use the \textit{Maximum Likelihood Estimation (MLE)} for the word inclusion and length control tasks; and use the unlikelihood training strategy~\cite{zhang-song-2022-discup} for the attribute control task.

It is worth noting that RMT shares the trait of training efficiency. Firstly, RMT efficiently reuses the output of the base CLM, which makes RMT requiring fewer pre-training datasets and expediting the pre-training process. Additionally, RMT is lightweight and is built on the top layer of the base CLM. Therefore, gradients do not need to propagate into the base CLM during the backpropagation process, which can save a significant amount of training time and GPU memory.

\section{Experiments}

\subsection{Word Inclusion Experiment}

Two different settings are experimented in this sub-section. The first one is the without-context setting, i.e., steering the CLM to generate a single sentence containing the keywords without the context. The second is the with-context generation setting, which requires the CLM to continue to generate target text under an extended context. The consideration of these two modes aims to test the CTG's flexibility, capable of controlling the generation of CLM in a position-independent way.

\begin{table*}[htbp]
\centering
\begin{tabular}{lccccc}
\toprule[1pt]
\multicolumn{1}{l}{\textbf{Method}} & \textbf{Cov ($\uparrow$)} & \textbf{PPL ($\downarrow$)} & \textbf{Self-Bleu ($\downarrow$)}  & \textbf{CS ($\uparrow$)} & \textbf{Fluency ($\uparrow$)}  \\ 
\midrule[1pt]
\multicolumn{4}{l}{CommonGen (without-context setting)} &  \\
% \midrule[1pt]
\midrule[0.5pt]
\text{POINT~\cite{zhang-etal-2020-pointer}}          & \textbf{98.0}    & 65.6          & 27.7          & 4.8           & 4.0          \\
\text{KG-BART~\cite{liu2021kg}}                      & 97.2             & 51.3          & 33.0          & 7.4           & 7.3           \\
\text{NeuroLogit~\cite{lu-etal-2022-neurologic}}    & 77.7             & \textbf{23.3}  & 56.2          & 4.6           & 4.5           \\

\text{NRP~\cite{carlsson-etal-2022-fine}}            & 93.0             & 42.3         & 28.4          & 6.1           & 7.0            \\
\text{Prompt+GPT2}                                   & 70.9             & 61.3         & 48.3          & 6.3           & 7.5            \\
\text{Prompt+ChatGPT}                                & 90.2             & 59.1         & -             & \textbf{7.9}  &\textbf{8.2}    \\

\text{RMT (w/o CL)}           & 93.1       & 56.4          & \textbf{20.9}                       & 6.5           & 6.7 \\ 

\text{RMT \textit{(CL=15)}}           & 93.9       & 60.3          & 22.5      &  -       & - \\ 
\text{RMT \textit{(CL=18)}}           & 93.7       & 47.9          & 23.7               & -      & - \\ 
\text{RMT \textit{(CL=20)}}           & 93.8       & 44.1          & 23.2               &  -       & - \\ 

\midrule[1pt]

\multicolumn{4}{l}{C2Gen (with-context setting)} & CS / Rel \\
% \midrule[1pt]
\midrule[0.5pt]
\text{Prompt+ChatGPT}&                              82.2                    & 46.7               & \textbf{5.2}     & \textbf{9.0 / 8.5}          & \textbf{8.8}    \\
$\text{NRP}^{*}$~\cite{carlsson-etal-2022-fine}          & 81.0                  & -      & -         & -               &-  \\

\text{RMT (w/o CL)}                                          & \textbf{91.8}         & \textbf{46.2}               & 5.8              & 7.1 / 8.3           & 6.4 \\ 

\bottomrule[1pt]

\end{tabular}
 
\caption{The experiment results on word inclusion. \textit{CL} represent the external control length setting in our experiment. We set the number of RMT block-layers to three ($M=3$). The result of $\text{NRP}^{*}$ in C2Gen is taken from what was reported in the original paper\protect\footnotemark. For the ChatGPT baseline in CommonGen, we test it on a subset of 500 samples, due to the problem of API access.}
\label{commongen}
\end{table*}
\footnotetext{It behaved not well in our actual test.}

\textbf{Experimental Setting.} Following the NRP~\cite{carlsson-etal-2022-fine}, we use the GPT2-large as the backbone CLM, and the training data for pre-training and fine-tuning also comes from Wikipedia. More details on training and inference are provided in Appendix~\ref{wordinc_details}. We test our approach on CommonGen~\cite{lin-etal-2020-commongen} (for without-context setting), and C2Gen~\cite{carlsson-etal-2022-fine}(for with-context setting). 

\textbf{Baselines.} As for the without-context setting, we firstly choose those methods that are specifically trained for lexical constraints tasks. They are KG-BART~\cite{liu2021kg}, which fine-tunes the whole parameters of BART while it is augmented by knowledge graph; and POINT~\cite{zhang-etal-2020-pointer}, which is an insertion-based method and iteratively injects words around the target words. Finally, we also employ a pure decode-time approaches for lexical constraint, namely NeuroLogic~\cite{lu-etal-2022-neurologic}, as a baselines. The general CTG  methods are also compared, including Non-Residual Prompt~\cite{carlsson-etal-2022-fine}(i.e., a recent state-of-the-art method using position-independent prompting model to guide the generation of GPT2-large), and approaches directly instructing the GPT2-large and ChatGPT to generate target sentences. As for the with-context setting, KG-BART and POINT could not be applied to this scene and thus they are removed. The test results of POINT, KG-BART, and NRP are adopted from the open source code\footnote{\url{https://github.com/FreddeFrallan/Non-Residual-Prompting}}.

\textbf{Evaluation.} Following the NRP~\cite{carlsson-etal-2022-fine},  We employ coverage (Cov), measuring the average coverage rate of the generated text with respect to the target words, to assess the controllability. Perplexity (PPL) is used to evaluate the fluency. We calculate PPL using an off-the-shelf GPT2-large model. Additionally, we utilize Self-Bleu-5 to evaluate the diversity of the generated text, with lower Self-Bleu values indicating greater syntactic diversity. Considering the non-intrusive setting (i.e., the CLM  is fixed and the task is conducted on the open-ended text generation setting), our experiments omit that task-specific metrics like BLEU, METEOR, CIDEr, and SPICE for CommonGen dataset. Additionally, we conduct a human evaluation to assess the conformity of the generated text to common sense (\textbf{CS}) and text \textbf{Fluency} from human perspective. For the with-context setting, we introduce an additional manual evaluation metric called Relevance (Rel), where human evaluators scored the relevance of the generated text to the given context. More detailed information can be found in the Appendix~\ref{human_evaluation}.

% which corresponds to around 16\% of the parameters of GPT2-large

\textbf{Result and Analysis.} As is shown in Table~\ref{commongen}, in the without-context setting, the coverage of our method significantly outperforms the vanilla prompt method (i.e., Prompt+GPT2) and pure decode-time approach (i.e, NeuroLogit) with the same decode algorithm using by RMT. It shows slight advantages over ChatGPT and NRP. Compared to task-specific models (i.e., POINT, KG-BART), the coverage of RMT is close to them, yet with lower PPL. It is worth noting that, in the with-context setting, the strong baselines, including ChatGPT and NRP, respectively suffer from a 12.0\% and 8.0\%  decline of control ability  compared to the no-context setting. However, RMT maintains the same-level of control ability with the no-context setting, and outperforms NRP and ChatGPT, which shows the superiority of RMT in that the control ability will not degrade with the context length.

In term of human evaluation in CommonGen, POINT is an insertion-based generation framework that naturally suffered from worse text quality; thus the result of RMT is much better than POINT  on terms of commonsense and fluency. Like NRP and Promp+GPT2, we all use frozen GPT-2-large as the backbone and do not introduce external knowledge. RMT achieves comparable results, which prove that the non-intrusive paradigm with residual learning would maintain the language model’s ability, thereby guaranteeing the text quality. Our approach falls behind KG-BART enhanced by knowledge graph and ChatGPT with hundred-billion level parameters. This is within exception. If inserting RMT into larger-scale CLM, e.g., GPT3, we believe the text quality of our approach will also be improved. As for C2Gen, RMT produces comparable results in contextual relevance with ChatGPT, but there is still a gap in common sense. \textit{More qualitative examples of different CTG approaches can be seen in Appendix~\ref{Qualitative_Examples}.}

% It is interesting RMT have a slightly higher PPL, Our method adds strict generation length control which further constrains the language model.

\subsection{Sentence Length Control Experiment}

\begin{table*}[htbp]
\renewcommand{\arraystretch}{1.4}

\resizebox{\textwidth}{!}{

\begin{tabular}{l|c|l}
% \begin{tabularx}{c|c|X}
% \hline
\toprule[1pt]
\textbf{Target Words} & \textbf{Control Length}   & \multicolumn{1}{c}{\textbf{RMT's Generated Text}}
\\ \hline
%% ****************************************************
\multirow{3}{*}{\color{red}{circle}, \color{blue}{sit}, \color{cyan}{talk}} & 13 & The group of people {\color{blue}{sitting}} around {\color{cyan}{talking}} in a {\color{red}{circle}} around them. 
\\
\cline{2-3}
& 17 & The woman is {\color{blue}{sitting}} in a {\color{red}{circle}} and {\color{cyan}{talking}} to a man in a white shirt.
\\
\cline{2-3}
& 25 & \makecell[l]{The group of people are {\color{blue}{sitting}} in a {\color{red}{circle}}, {\color{cyan}{talking}} about what they’re \\ going to do with their lives.} 
\\
\hline
%% ****************************************************
\multirow{3}{*}{\color{red}{drink}, \color{blue}{sit}, \color{cyan}{table}, \color{magenta}{wine}} & 13 & The man is {\color{blue}{sitting}} on a {\color{cyan}{table}} and {\color{red}{drinking}} {\color{magenta}{wine}} from a glass.
\\
\cline{2-3}
& 17 & The man is {\color{blue}{sitting}} on a  {\color{cyan}{table}} with a {\color{red}{drink}} in his hand and {\color{red}{drinking}} {\color{magenta}{wine}}.
\\
\cline{2-3}
& 25 & \makecell[l]{The man is {\color{blue}{sitting}} at a {\color{cyan}{table}} with a {\color{red}{drink}} in his hand, while another man \\ is {\color{red}{drinking}}  {\color{magenta}{wine}} from a glass.}
\\
%% ****************************************************
\bottomrule[1pt]
\end{tabular}
% \end{tabularx}
}
\vspace{0.1cm}
\caption{The examples of generated text under different control length. The generated length is counted with GPT2's Tokenizer, thus it will make a little difference with the number of actual words.}
\label{control_length}
\end{table*}
%%% Original
%%% ************************************************************

We test the length-control ability of RMT, i.e., instructing the CLM to generate a sentence which satisfies the word inclusion objective under an exact number of words. This setting increases the challenge of word inclusion task, effectively approving the fine-gained control ability. We observe that RMT can plan accordingly by the instructions of the required token numbers and target words. 

% The control instructions are designed as follows: \textit{Include: $\text{word}_1$, $\text{word}_2$, ..., $\text{word}_n$; Context Length: $m$; Target Length: $n$.}

% The control instructions are designed as follows:
% \textcolor{brown}{\textit{Include: word1, word2...word\_n; Context Length: m; Target Length:n.}} \\

We report control performance for $CL=15, 18, 20$. As shown in Table~\ref{commongen}, RMT can maintain same level word coverage under different control-length.  The longer the target length, the lower the generated text's PPL. This result is quite intuitive since it is harder to produce fluent text within the shorter required length.

%% %%%%%%%%%%%%%%%%%%%%%%%%%%%%%%%%%%%%%%%
%% Control Length Figure.
\begin{figure}
\vspace{-1.1em}
\centering
\includegraphics[width=0.4\textwidth]{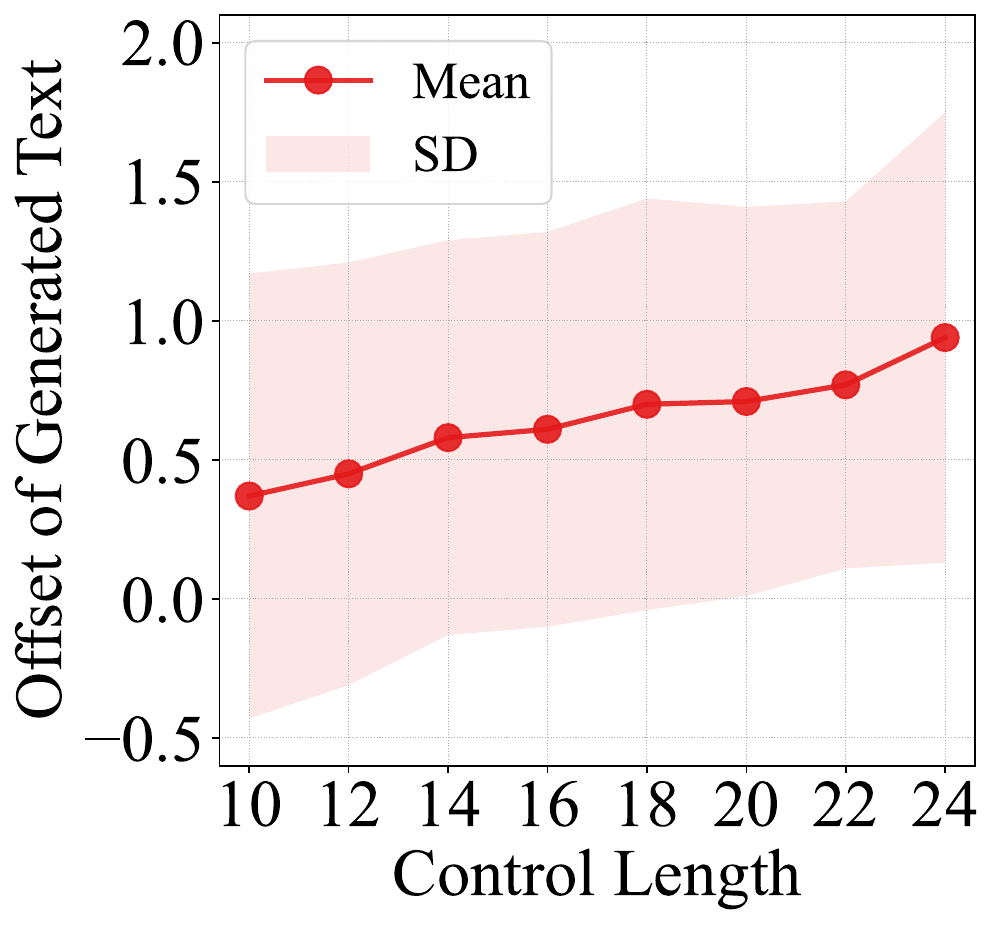}
\caption{\label{fig:length} Control length and control performance.}
% \vspace{-2.0em}
\end{figure}
%% %%%%%%%%%%%%%%%%%%%%%%%%%%%%%%%%%%%%%%%

To quantitatively measure the controllability of sentence length, we conduct tests on the CommonGen validation dataset to analyze the control effect of our proposed method. Some examples are presented in Table~\ref{control_length}, which demonstrate that RMT effectively steers the CLM to generate texts with specific keywords and required sentence lengths. Furthermore, we calculated the mean offset (Mean) and the standard deviation (SD) for different control lengths. The results are visualized in Figure~\ref{fig:length}. We observe that the offset and standard deviation both remain within one, indicating that RMT successfully achieves the sentence generation with desired sentence lengths as instructed.

%%% ************************************************************
%%% SunSi Revise

\subsection{Attribute Control Experiment}
% In theory, the structural features of RMTS can be applied to any controllable generation scenario; thus we also conduct the sentiment control experiment to test RMT's performance on the term of attribute control to approve the versatility of our approach further.

\textbf{Setting and Baselines.} We follow DisCup~\cite{zhang-song-2022-discup}, and use a discriminator-involved training strategy to fine-tune RMT, and the checkpoint of attribute-classifier comes from the open source code\footnote{ \url{https://github.com/littlehacker26/Discriminator-Cooperative-Unlikelihood-Prompt-Tuning} \label{discup_code}}. During the training, we optimize RMT using two different strategies, one is to use top-k=90 algorithm to select re-rank candidate tokens and the other setting uses top-p=0.95 to select candidate tokens. The top-k setting increases the depth of token sampling, thus the diversity and control ability of the model will increase, yet with the burden of decreasing text fluency. The top-p keeps the re-ranked tokens within the distribution of base-CLM, which could potentially achieve higher fluency but lower text quality and control ability. More details could be seen in ~\cite{zhang-song-2022-discup}. Different from DisCup, which optimizes unique prompt for every attribute, we use the RMT module to directly encode different attribute instructions uniformly. GPT2-large is used as the backbone CLM, and the training data is the widely used Stanford Sentiment Tree (SST-5)~\cite{socher-etal-2013-recursive} collected from movie reviews. We follow the commonly used setting in previous work~\cite{gedi, zhang-song-2022-discup, lu2022quark}. Specifically we use 5K neutral prompts, 2.5K positive prompts, and 2.5K negative prompts as test dataset (provided by DEXPERT ~\cite{liu-etal-2021-dexperts}), and every CTG approach generates another 20 continuations based on given prompts, to achieve sentiment (i.e., positive and negative) controllable text generation.  We collect the primary baselines reported by DisCup~\cite{zhang-song-2022-discup} and DEXPERT~\cite{liu-etal-2021-dexperts}, including the decode-time approaches with GPT2-large as base-CLM (i.e., PPLM~\cite{pplm},GEDI~\cite{gedi},and DEXPERT~\cite{liu-etal-2021-dexperts}), and the training approaches (i.e., CTRL~\cite{ctrl} and DisCup~\cite{zhang-song-2022-discup}).

% \footnote{ \url{https://huggingface.co/distilbert-base-uncased-finetuned-sst-2-english} \label{sentiment_classfier}}

\textbf{Evaluation.} Following previous work~\cite{gedi, zhang-song-2022-discup, lu2022quark}, we use an external sentiment classifier provided by \texttt{Huggingface.co} to classify the generated texts, and get sentiment control accuracy (i.e., Correctness). PPL and Dist-1/2/3 are reported to test their fluency and diversity, respectively. For human evaluation, we introduce \textit{Relevance}, i.e., how the text conforms to required sentiment; \textit{Topicality}, i.e., how the generated continuations are context-consistent with the given prompts, and \textit{Fluency}, i.e., the text's fluency evaluated from the human perspective. More details can be seen in Appendix~\ref{human_evaluation}.

\begin{table*}[t]

% Correctness represents the proportion of generated sentences satisfying the target sentiment. We use the PPL measured by GPT2-large to evaluate the texts' fluency, and Dist-1/2/3 to measure the diversity of text.

\resizebox{\linewidth}{!}{
\begin{tabular}{clccccc}
\toprule[1pt]
&                                    & \multicolumn{3}{c}{\textbf{Correctness} ($\uparrow$)}          & \multicolumn{1}{l}{\textbf{Fluency} ($\downarrow$)} & \textbf{Diversity} ($\uparrow$)  
                                            \\
\multirow{-2}{*}{\textbf{Target Sentiment}} & \multirow{-2}{*}{\textbf{Method}}  &  Positive & Neutral & Negative    & PPL                   & Dist-1 / Dist-2 / Dist-3                  
\\
\midrule[0.5pt] 
\multirow{6}{*}{Positive}
& $\text{PPLM~\cite{pplm}}$ & \multicolumn{1}{l}{} & 52.68            & 8.72                 & 113.54          
                                            & 0.39 / 0.83 / 0.89                                                                              \\ 
                                            & $\text{CTRL~\cite{ctrl}} $                            & \multicolumn{1}{l}{} & 77.24            & 18.88                & 48.24                                & 0.13 / 0.53 / 0.79                                                                         \\
                                            & $\text{GEDI~\cite{gedi}}$                               & \multicolumn{1}{l}{} & 86.01            & 26.80                 & 123.56                           & 0.20 / 0.66 / 0.85                                                                           \\
                                            & $\text{DEXPERT~\cite{liu-etal-2021-dexperts}} $         & \multicolumn{1}{l}{} & 94.46            & 36.42                 & 60.64                            & 0.18 / 0.63 / 0.84                                                                            \\

& $\text{DisCup~\cite{zhang-song-2022-discup}}$           & \multicolumn{1}{l}{} & 94.20   & 60.40       & 46.6            
& 0.14 / 0.51 / 0.78  \\
                                            & $\text{RMT+top-k}$ & \multicolumn{1}{l}{} &  \textbf{97.62}       & \textbf{67.20}    &   46.0  & 0.14 / 0.56 / 0.79  \\
                                            & $\text{RMT+top-p}$ & \multicolumn{1}{l}{} &  94.50             &  42.60      & \textbf{17.3}  & 0.13 / 0.45 / 0.65  

\\ \hline
\multirow{6}{*}{Negative}
                                            & $\text{PPLM~\cite{pplm}} $                                & 10.26                & 60.95            & \multicolumn{1}{l}{} & 122.41                           & 0.40 / 0.83 / 0.90                                                                     \\
                                            & $\text{CTRL~\cite{ctrl}} $                              & 20.95                & 62.37            & \multicolumn{1}{l}{} &  45.27                              & 0.13 / 0.51 / 0.78                                                                           \\
                                            & $\text{GEDI~\cite{gedi}} $                               & 60.43                & 91.27            & \multicolumn{1}{l}{} & 138.93                            & 0.19 / 0.66 / 0.86                                                                           \\
                                            & $\text{DEXPERT~\cite{liu-etal-2021-dexperts}}$                    & 64.01                & \textbf{96.23}   & \multicolumn{1}{l}{} & 67.12                      & 0.20 / 0.64 / 0.83    

                                            \\
                                            
& $\text{DisCup~\cite{zhang-song-2022-discup}}$          & 62.80       & 91.40            & \multicolumn{1}{l}{} & 47.90    & 0.13 / 0.50 / 0.77                                                                              \\

                                            & $\text{RMT+top-k}$          & \textbf{77.16}       & 95.92            & \multicolumn{1}{l}{} & 49.15    & 0.15 / 0.60 / 0.82  \\

                                            & $\text{RMT+top-p}$        & 56.4      & 92.7            & \multicolumn{1}{l}{} & \textbf{19.0}    & 0.13 / 0.48 / 0.70  \\ 
                                            
                                    \bottomrule[1pt]
\end{tabular}

}
\caption{The experimental results of sentiment controllable text generation. Among the table, $\uparrow$ indicates that the higher corresponding value is better, and $\downarrow$ is the opposite.}

\label{automatic_seniment_result}
% \vspace{-0.5em}
\end{table*}

\begin{table} % {l}{8cm}
\resizebox{\linewidth}{!}{
\begin{tabular}{lccc}
\toprule[1pt]
\multicolumn{1}{l}{\textbf{Method}} & \textbf{Relevance ($\uparrow$)} & \textbf{Fluency ($\uparrow$)} & \textbf{Topicality ($\uparrow$)} \\ 
\midrule[1pt]
$\text{DisCup}$~\cite{zhang-song-2022-discup}            & 6.3            & 6.6               & 6.5                 \\
$\text{DEXPERT}$~\cite{liu-etal-2021-dexperts}           & 6.2            & \textbf{7.1}      & \textbf{7.4}         \\
$\text{RMT+top-k}$                                       & \textbf{7.8}   &7.0                & 7.2        \\ 
\bottomrule[1pt]
\end{tabular}}

% \caption{The human evaluation results on sentiment control experiment.}
\caption{Human evaluation results of sentiment control. As for each metric, $\uparrow$ indicates that the higher corresponding value is better, and $\downarrow$ is the opposite. }
% \vspace{-1.0em}
\label{human_evaluate}
\end{table}

% Both methods have their merits.
\textbf{Result and Analysis.} As shown in Table~\ref{automatic_seniment_result}, RMT in top-k setting demonstrates a superior control performance compared to all baselines, while maintaining comparable text quality and diversity. And RMT in the top-p setting shows better text quality yet weaker control performance. The closest performers to our approach are DEXPERT and DisCup. However, DEXPERT requires fine-tuning an external pair of CLM, resulting in tuning two time more parameters of the base CLM. In contrast, RMT is parameter-efficient, requiring tuning relevant parameters which are only around 16\% of the base CLM's parameters. Regarding DisCup, both approaches employ the same loss function. However, DisCup optimizes different continuous prompts for each attribute, while we utilize the RMT to steer the CLM using residual learning. DisCup requires fine-tuning fewer parameters, while RMT is completely decoupled from the CLM and controls different attributes using a unified model, providing a greater flexibility and a more powerful control capability.

The human evaluation result presented in Table~\ref{human_evaluate} indicates that DEXPERT performs slightly better in terms of fluency and contextual relevance, but exhibits lower control ability, and RMT excels in attribute relevance. We speculate that the use of a classifier-based objective in RMT and DisCup might result in overly strong control, leading to the generation of abrupt turning points that may compromise fluency and topicality to some extent. DisCup performs relatively poorly, as too few control parameters limit its expression ability. \textit{More detailed examples are presented in Appendix~\ref{Qualitative_Examples}.}

\subsection{Further Analysis}

\begin{figure} %{r}{4.0cm}
\centering
% \vspace{-1.5em}
\includegraphics[width=0.3\textwidth]{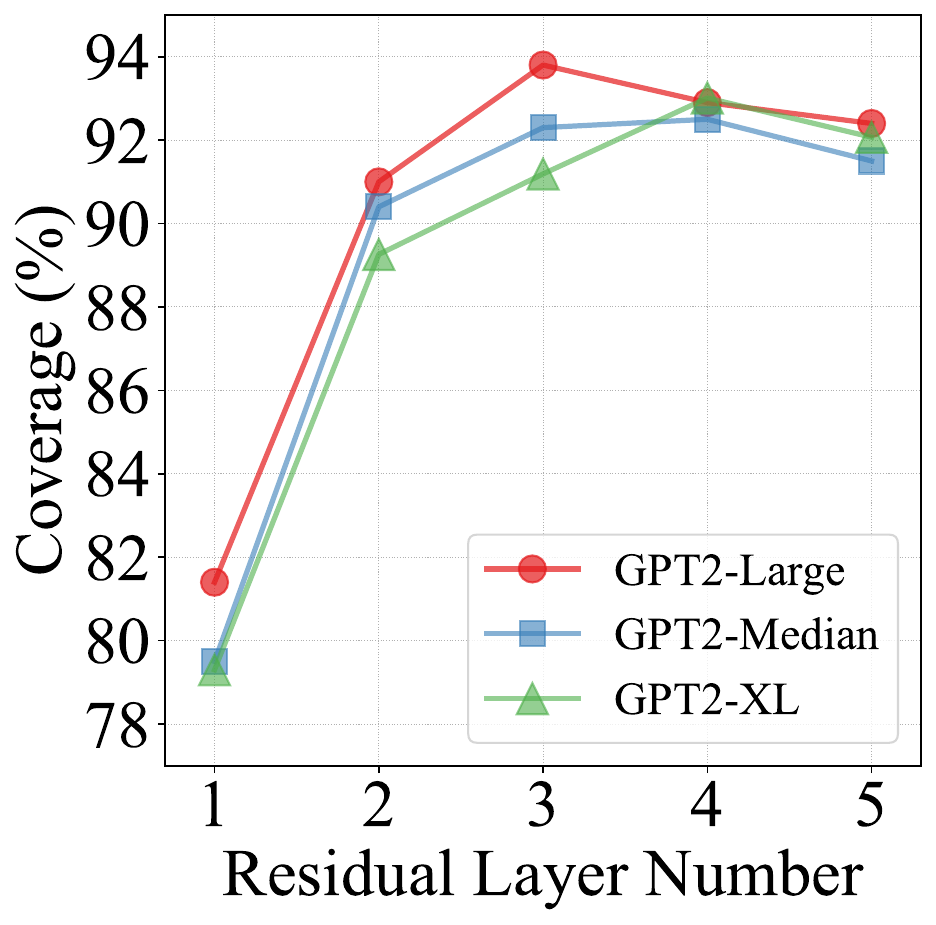}
% \caption{\label{fig:cov_ppl} Influence of the number of block layer on control performance, the result is tested under CommonGen dataset.}
% \vspace{-0.5em}
\caption{\label{fig:cov_ppl} RMT block layers and control performance. RMT could achieve a consistently high and stable control performance for CLMs with different sizes. About three block layers are sufficient for optimal performance.}
% \vspace{-0.5em}
\end{figure}
\textbf{Block Layers and base-CLM size.} In order to test the scalability over the number of block layers and base-clm,  we investigate the influence of the number of block layers on the RMT's control performance over different size of base-CLM models. The results shown in Figure~\ref{fig:cov_ppl}  indicate that a value of $M=3$ is deemed appropriate for achieving effective word inclusion and sentence length control. This suggests that RMT has the advantage of being parameter-efficient over the similar approaches like NRP~\cite{carlsson-etal-2022-fine} that necessitates an external prompting model of the same size as the original CLM. Moreover, RMT achieves a consistently high and stable control performance for CLMs with different sizes, which indicates that RMT is potentially applied to various size of CLM models.

%% %%%%%%%%%%%%%%%%%%%%%%%%%%%%%%%%%%%%%%%
%% Coverage & PPL Figure.

%% %%%%%%%%%%%%%%%%%%%%%%%%%%%%%%%%%%%%%%%

% \vspace{-4.0em}

\begin{table} %{l}{4.5cm}
\centering
% \vspace{-2em}
% \resizebox{\linewidth}{!}{
\begin{tabular}{lcc}
\\ \toprule  
\textbf{Models} & \textbf{Cov (\%)}  \\\midrule
RMT  &\textbf{93.8} \\
\textit{w/o causal Self-Att} &73.2 \\  
\textit{w/o causal CLM-Att} &75.1 \\
\textit{w/o Pre-training} &74.5 \\   \bottomrule
\end{tabular}
% }
\caption{Ablation study of RMT.}\label{wrap-tab:1}
\label{ablation}
% \vspace{-1em}
\end{table}

\textbf{Ablation Study.} To assess the effectiveness of the Residual Memory Transformer (RMT), we conduct an ablation study. The study involves three settings: removing the causal self-attention, excluding the causal CLM attention, and evaluating RMT's performance without pre-training. The results are presented in Table~\ref{ablation}. Interestingly, under all three settings, we observe a significant decline in performance. This finding serves as compelling evidence that every component in RMT, as well as the pre-training stage, is crucial and indispensable for achieving optimal results.

\textbf{Generation Efficiency.} RMT is a lightweight plugin, allowing for efficient inference speeds comparable to the original CLM. Table~\ref{inference_speed} demonstrates that RMT outperforms typical CTG approaches in term of efficient generation speed, approaching the speed of the pure CLM (GPT2-large).

\section{Related Work}

% In recent years, numerous CTG approaches based on pre-trained language models (PLMs) have emerged.
As summarized by~\cite{ctg_survey}, CTG approaches could be divided into three categories, i.e., retraining/refactoring, fine-tuning, and post-process. The first two categories refer to the training approaches, which aim to fine-tune (e.g., reinforcement learning is used to fine-tune pre-trained language model(PLM)~\cite{ouyang2022training,lu2022quark}, optimizing continuous prompts~\cite{tailor_2022, zhang-song-2022-discup}, and prompting model~\cite{carlsson-etal-2022-fine} to steer text generation) or retrain ~\cite{chan2021cocon, ctrl, DART} a PLM to generate texts that meet the desired control condition. These methods have shown significant performance improvements in the field. However, with the increasing size of PLM, they have become resource-intensive to fine-tune or retrain. As a result, post-process approaches have become more popular in the research community.

\begin{table} %{r}{5.0cm}
\centering
% \vspace{-1.5em}
\resizebox{\linewidth}{!}{
\begin{tabular}{lc}
\toprule[1pt]
\multicolumn{1}{l}{\textbf{Method}} & \textbf{Time (second)}
\\ 
\midrule[0.5pt]
PPLM~\cite{pplm}                                & 37.39                            \\
$\text{Mix-Match}^{\ast}$~\cite{mireshghallah-etal-2022-mix}                       & 33.5                             \\
$\text{COLD}^{\ast}$~\cite{qin2022cold}                                & 33.6                             \\
DEXPERT~\cite{liu-etal-2021-dexperts}                             & 2.54                             \\
DisCup~\cite{zhang-song-2022-discup}                              & 0.94                             \\ 
GPT2-large                         & 0.78                             \\
RMT                          & 0.88                             \\ 

\bottomrule[1pt]
\end{tabular}
}

\caption{The generation efficiency of RMT. The cost (time) for generating 20 tokens based on GPT2-large. The results of methods marked with $\ast$ are reported from~\cite{qin2022cold}.}
\label{inference_speed}
\vspace{-2em}
\end{table}

The post-process methods are dedicated to guiding language model toward desired texts in the decode-time stage using an auxiliary module. Achieving the goal of generating attribute-specific texts, PPLM~\cite{pplm} leverages a simple attribute classifier to update the head hidden layer of LM by gradient feedback. Then, Diffusion-LM~\cite{li2022diffusion} combines Diffusion-LM and classifier-guided gradient updates to the continuous sequence of latent variables, achieving plug-and-play controllable generation. COLD~\cite{qin2022cold} proposes a novel decoding framework, by combining energy-based model and gradient-based sampling. Fudge~\cite{yang-klein-2021-fudge} uses the discriminator that contains future attribute information to re-rank the token distribution produced by a frozen GPT2 model. NeuroLogits~\cite{lu-etal-2022-neurologic,lu-etal-2021-neurologic} incorporates the lexical constraints into the decode-time algorithm. In order to accelerate the generation process, GeDi~\cite{gedi} and DEXPERT~\cite{liu-etal-2021-dexperts}  train another smaller language model as generative discriminators to guide the generation from a base GPT2. Plug-and-Blend~\cite{lin2021plug} extends GeDi to controllable story generation by introducing a planner module.
% \cite{meng2022controllable} propose to decompose the oracle into token-level guidance to steer the base model in the generation, as such, to achieve the control of sentence-level attributes. 
Moreover, controlling multiple control elements is also expolred~\cite{kumar2021controlled, tailor_2022}. 

Most decoder-time approaches either solely intervene with the LM during the token selection phase of generation, lacking planning capabilities, or require multiple iterations, resulting in excessive generation times. Our approach shares a plug-and-play trait with decoder-time approaches. However, the key distinction is that we use a lightweight residual model to integrates control information and multi-level contextual streams from the CLM, enabling fine-gained content planning and efficient text generation.
 
\section{Discussion and Future Work}

RMT still suffer some the limitations. (1) Challenges in applying RMT to close-ended CLMs. Presently, the application of RMT needs to obtain the last hidden states or the logits of CLM, thus applying RMT to some commercial CLMs, e,g., GPT-4, still faces challenges. This is also a common problem of all plugin-style CTG.  (2) RMT does not focus on commonsense, may result in generating texts which are not confirmed to commonsense sometimes. This issue could be potentially relieved by introducing  external knowledge graph in the future work.

% Due to limited computer resources and open source issues, we have not tried to apply RMT to larger CLM models (e.g., GPT-3~\cite{gpt3}, PaLM~\cite{chowdhery2022palm}, and LLaMA~\cite{touvron2023llama}, etc.) in our experiments. However, our approach is theoretically model-agnostic; hence, we eagerly look forward to  applying RMT to those bigger models in the future.

Up to now, it is still challenging to avoid the factual errors that appeared in the generated text for large-scale causal language models, which RMT does not address either. A promising future work is to combine RMT and information retrieval systems to enhance the factual accuracy of those generative models. Moreover, RMT could also be used to encode personal profiles and build personalized chatbots, or fusion with image information so as to be applied to multi-modal scenes.

\section{Conclusion}
In this paper, we have proposed a new CTG alternative,  which leverages a residual model to steer the CLM to generate desired text noninvasively. Additionally, we propose a residual memory transformer, a novel encoder-decoder architecture, to fuse the raw contextual text, generative stream of CLM, and control information in a shot, thus better collaborating and controlling the generation of CLM. The experiments show that RMT exhibits better performance in flexibility, control granularity, and efficiency, making it a compelling solution for controllable text generation.


% Use unnumbered third level headings for the acknowledgments. All acknowledgments
% go at the end of the paper. Do not include acknowledgments in the anonymized
% submission, only in the final paper.

% \section*{References}

% \bibliographystyle{plain} 
\bibliography{sample-base}

\appendix
\newpage

\section*{Appendices}
In order to enhance the reproducibility of the experimental results presented in our paper and provide additional supporting details for the conclusions outlined on the main page, we have included supplementary materials in this section.

%These materials consist of experimental details,human evaluation results, generation efficiency analysis, and generation examples.

\appendix

\section{Experimental Details}

In our paper, all the experiments are conducted on a single NVIDIA A6000 GPU with 48GB of memory. We utilize the PyTorch deep learning framework as our implementation tool and employ the HuggingFace Transformers package to load a pre-trained GPT model. To implement the Residual Memory Transformer (RMT), we utilize the Transformer package in PyTorch-1.14.0, and the number of multi-head attention is set to 8. 

\subsection{Pre-training} 
\label{pretraining_details}

\textbf{Data Details.} The pre-training stage is consistently applied across all experimental settings in our study. For this stage, we collect a dataset of 4 million sentences sampled from the Wikipedia corpus to serve as the training data. To optimize computational resources, we filtered out sentences that exceeded a length of 200 words. These samples are then noised using the following methods: 

\begin{itemize}
  \item [(1)] Token Masking: We randomly mask tokens with a special token. The proportion of masked tokens in a sample is 20\%.     
  \item [(2)] Token Deletion: We randomly delete tokens from the text. The proportion of deleted tokens in a sample is 20\%.
  \item [(3)] Span Replacing: Text spans of various lengths are randomly replaced with special tokens. Each span is replaced with a single special token. The proportion of replaced span tokens in a sample is 20\%.
  \item [(4)] Text Rotation: A token is randomly selected, and the text is rotated around it. This means the text began with the selected token and ended with its previous token. 
\end{itemize}

% (1) Token Masking: We randomly replace tokens with a special token. The proportion of masked tokens in a sample is 20\%. \\
% (2) Token Deletion: We randomly delete tokens from the text. The proportion of deleted tokens in a sample is 20\%. \\
% (3) Span Replacing: Text spans of various lengths are randomly replaced with special tokens. Each span is replaced with a single special token. The proportion of masked span tokens in a sample is 20\%. \\
% (4) Text Rotation: A token is randomly selected, and the text is rotated around it. This means the text began with the selected token and ended with its previous token. 

In the total dataset, the ratios of \textit{Token Masking}, \textit{Token Deletion}, \textit{Span Replacing}, and \textit{Text Rotation} are 50\%, 16.7\%, 16.7\%, and 16.7\%, respectively. Since RMT shares GPT's tokenizer, which does not reserve a special token, we select the word ``xxx'' with no special meaning as the special token.
we did not specifically tune the hyperparameters related to the ratios of different denoised tasks during pretraining. Instead, we followed BART's [12] approach and selected the most important tasks, such as Text Infilling (included by Token Masking in RMT), to constitute the largest proportion of the task. The remaining tasks were divided equally among the other three tasks. In our experiments, we observed that the specific choice of pretraining tasks did not significantly impact the results in CTG. For instance, when we used Token Masking as the only pretraining task, the control performance on CommonGen reached 92.2\%. By adding the other three tasks, the performance improved slightly to 93.9\%.

\textbf{Training Details.} During the pre-training process, we set the following parameters for model training: the batch size of 64, learning rate of 5e-5, random seed of 42, and pre-training epoch of 1. We employ the AdamW as the optimizer; the pre-training of the RMT is conducted on a single NVIDIA A6000 GPU. Since the backpropagation process does not need to propagate into the Casual Language Model (CLM), the training of the model was efficient. As a result, the entire pre-training stage is completed in approximately 30 hours.

\subsection{Word Inclusion}
\label{wordinc_details}

\textbf{Baselines.} We closely follow the methodology of previous work, specifically the Non-Residual Prompt (NRP)~\cite{carlsson-etal-2022-fine}. To establish a reliable reference point for the CommonGen dataset, we rely on the generation results of the baselines provided in the NRP paper. And for the NeuroLogic baseline, in order to keep the comparability, we use the same decoding algorithm setting with RMT, the beam search size is 4 and max generation length is set to 32. For the C2Gen dataset, we utilize the checkpoint provided by NRP, to generate the responding texts. However, the result shows NRP could not work well. As a result, we give up using NRP as a baseline in the C2Gen dataset.

Regarding the ChatGPT baseline, we employ prompts specifically designed for generating results on the CommonGen dataset. The specific prompt is as follows: \\
\textcolor{brown}{\textit{Continue to write a sentence, and include [$\text{word}_1$, $\text{word}_2$, ..., $\text{word}_n$], the sentence length should be within 15 words.}} \\
When it comes to the C2Gen dataset, the prompt instruction of ChatGPT is:\\
\textcolor{brown}{\textit{Continue to write a sentence, and include [$\text{word}_1$, $\text{word}_2$, ..., $\text{word}_n$], the sentence length should be within 15 words. Context: [context text].}}

\textbf{Details of RMT.} In our approach, we randomly select training data from the Wikipedia corpus and fuse the training data of CommonGen. Specifically, for the setting involving without-context word inclusion and sentence length control, we gather a total of 158,343 samples. To ensure manageable sentence lengths, we impose a limit of 32 tokens per sentence. To choose the appropriate checkpoint during training, we utilize the CommonGen validation dataset and selected the checkpoint based on coverage metrics. For the with-context setting, we collect 132,170 samples, each with a sentence length constraint of 128 tokens, and constructed a validation dataset to choose the checkpoint.

\begin{table} %{l}{4.8cm}
% \vspace{-0.4cm}
% \resizebox{\linewidth}{!}{
\begin{tabular}{lc}
\toprule[1pt]
\multicolumn{1}{l}{\textbf{Name}} & \textbf{Value} \\ 
\midrule[0.5pt]
num\_beams                   & 4                            \\
top\_p                       & 0.7                           \\
repetition\_penalty          & 1.25                           \\
no\_repeat\_ngram\_size      & 3                             \\
\bottomrule[1pt]
\end{tabular}
% }
% \vspace{0.2cm}
\caption{Hyper-parameters for text generation in the inference stage.}
\label{app:hyper_paramer}
% \end{table}
% \vspace{-0.4cm}
\end{table}

Every sample is composed of two parts: context and target. We use spacy library\footnote{\url{https://spacy.io/}} to extract the keywords in the target text, of which are tagged with \texttt{VERB}, \texttt{NOUN}. The number of keywords in each sentence is restricted to the range from 3 to 5. The context length, target length, and keywords are used as control instructions to steer the generation process of the CLM. All above training data will be released and open source, once we have cleaned up our project.

During the fine-tuning stage, the learning rate of the AdamW optimizer is set to 1e-4, and the max epoch and batch size are 6 and 64, respectively. During the inference stage, For the control task without context (i.e., CommonGen dataset), we use the specific word ``The'' as context; the core parameters of the decoding algorithm are shown in Table~\ref{app:hyper_paramer}.

% % \begin{table}[htbp]
% % \centering
% \begin{wraptable}{l}{5cm}
% \resizebox{\linewidth}{!}{
% \begin{tabular}{lc}
% \toprule[1pt]
% \multicolumn{1}{l}{\textbf{Name}} & \textbf{Value} \\ \midrule[1pt]
% num\_beams                   & 4                            \\
% top\_p                       & 0.7                           \\
% repetition\_penalty          & 1.25                           \\
% no\_repeat\_ngram\_size      & 3                             \\
% \bottomrule[1pt]

% \end{tabular}}
% \vspace{0.2cm}
% \caption{Hyper-parameters for text generation in the inference stage.}
% \label{app:hyper_paramer}
% % \end{table}
% \end{wraptable}

\subsection{Sentiment Control}
\textbf{Baselines.} In this part, we build upon the baselines established by DisCup~\cite{zhang-song-2022-discup} and DEXPERT~\cite{liu-etal-2021-dexperts}, adopting their settings, checkpoints, and generated results. The results of baselines, including PPLM~\cite{pplm}, CTRL~\cite{ctrl}, GEDI~\cite{gedi}, and DEXPERT, are calculated from the generated texts provided DEXPERT~\cite{liu-etal-2021-dexperts}. As for DisCup, we use the checkpoint from the authors, the depth of re-ranked candidate token is set to top-k=9 to generate continuations using the prompts in test datasets. 

\textbf{Details of RMT.} In our own approach, during the training stage, we incorporate the discriminator-involved loss introduced by DisCup and utilize the attribute classifier provided by their open-source code. The instructions for sentiment control are ``\textit{The sentiment: positive}'' and ``\textit{The sentiment: negative}'', respectively. We set the size of re-ranked candidate tokens to 90, and choose a low temperature value of 0.005 and a learning rate of 2e-4. During the inference stage, we employ the same generation parameters as those used in the word inclusion experiments described in Table~\ref{app:hyper_paramer}. 

\section{CTG Approaches Comparison}
\label{CTG_Approaches_Comparison}

\subsection{CTG Comparison}

%% *******************************************
%% Table
%% *******************************************
\begin{table*}
\centering
% {r}{8.5cm}
% \vspace{-0.35cm}
\resizebox{\linewidth}{!}{
\begin{tabular}{lccc}
\toprule[1pt]
\multicolumn{1}{l}{\textbf{CTG Approach}} 
& \multicolumn{1}{l}{\textbf{Fine-gained Control}} 
& \multicolumn{1}{l}{\textbf{Efficient Generation}} 
& \multicolumn{1}{l}{\textbf{Non-intursive}} 
\\ \midrule[0.5pt]
%% ****************************************
PPLM~\cite{pplm}  
& {\color{Red}\ding{56}}
& {\color{Red}\ding{56}} 
& {\color{Green}\ding{52}} 
\\
%% ****************************************
CTRL~\cite{ctrl} 
& {\color{Green}\ding{52}} 
& {\color{Green}\ding{52}} 
& {\color{Red}\ding{56}}
\\
%% ****************************************
NeuroLogit~\cite{lu-etal-2022-neurologic} 
& {\color{Red}\ding{56}}
& {\color{Red}\ding{56}}
& {\color{Green}\ding{52}} 
\\
%% ****************************************
GEDI~\cite{gedi} 
& {\color{Red}\ding{56}}
& {\color{Green}\ding{52}}
& {\color{Green}\ding{52}} 
\\
%% ****************************************
DEXPERT~\cite{liu-etal-2021-dexperts}
& {\color{Red}\ding{56}} 
& {\color{Green}\ding{52}}
& {\color{Green}\ding{52}}                                        
\\
%% ****************************************
FUDGE~\cite{yang-klein-2021-fudge}                                 
& {\color{Red}\ding{56}}                                    
& {\color{Red}\ding{56}}                                 
& {\color{Green}\ding{52}}                                        
\\
%% ****************************************
% NADO~\cite{meng2022controllable}                                    
% & {\color{Red}\ding{56}}                                   
% & {\color{Green}\ding{52}}                                 
% & {\color{Red}\ding{56}}                                        
% \\
%% ****************************************
COLD~\cite{qin2022cold}                                    
& {\color{Green}{\color{Green}\ding{52}}}                      
& {\color{Red}\ding{56}}                                 
& {\color{Green}\ding{52}}                                        
\\
%% ****************************************
Diffsion-LM~\cite{li2022diffusion}                             
& {\color{Green}\ding{52}}                                   
& {\color{Red}\ding{56}}                                 
& {\color{Green}\ding{52}}                                        
\\
%% ****************************************
Mix\&Match~\cite{mireshghallah-etal-2022-mix}       
& {\color{Red}\ding{56}}                                   
& {\color{Red}\ding{56}}                                 
& {\color{Green}\ding{52}}                                        
\\
%% ****************************************
CoCon~\cite{chan2021cocon}                   
& {\color{Green}\ding{52}}                                   
& {\color{Green}\ding{52}}                                  
& {\color{Red}\ding{56}}                                         
\\
%% ****************************************
DisCup~\cite{zhang-song-2022-discup}                                  
& {\color{Red}\ding{56}}                                    
& {\color{Green}\ding{52}}                                 
& {\color{Red}\ding{56}}                                         
\\
%% ****************************************
NRP~\cite{carlsson-etal-2022-fine} 
& {\color{Green}\ding{52}}                                   
& {\color{Green}\ding{52}}                                  
& {\color{Red}\textbf{?}}                                        
\\
%% ****************************************
RMT(ours)                               
& {\color{Green}\ding{52}}                                   
& {\color{Green}\ding{52}}                                  
& {\color{Green}\ding{52}}                                        
\\ 
%% ****************************************
\bottomrule[1pt]
\end{tabular}
}
\vspace{0.1cm}
\caption{An overview of the key features in CTG approaches.}
\label{ctg_compare}
% \vspace{-0.4cm}
\end{table*}

We select some notable works in controllable text generation (CTG) and compare their characteristics. The results are shown in Table~\ref{ctg_compare}. We compare them according to three key features: \textbf{\textit{(1) Fine-grained Control}} suggests whether the method can potentially be applied to control various elements, such as emotion, topic, toxicity avoidance, word constraints, and structural control (e.g., personal profile, syntax tree). It indicates the method's ability to handle different control aspects effectively. \textbf{\textit{(2) Efficient Generation}} evaluates whether the time-cost of text generation using the method is comparable to that of the original language model (LM). \textbf{\textit{(3) Non-intrusive}} considers whether the method can operate without disturbing the language model's original textual stream, mitigating the harms of LMs.  A non-intrusive approach ensures flexibility and practicality in controlling LMs. Specifically, the non-intrusive fashion of RMT, which decouples the control module from the base-CLM, offers numerous exciting possibilities. One significant advantage is that we no longer need to fine-tune the base CLM for specific control tasks, avoiding any adverse impact on its generalization ability (i.e.,keeping the CLM from becoming task-specific model). Moreover, this design enables us to dynamically adjust the fine-grained control requirements during the text generation process at any time. For instance, we can flexibly control the sentiment or topic of individual sentences/paragraphs within a story by incorporating RMT into a story-writing CLM. Additionally, we can enhance a well-trained CLM, such as ChatGPT, with external grounded documents, making it possible to determine whether external control information should be Intervened through an automatic selection model (i.e., switching b/w controlled and free-style generation).

The analysis suggests that controlling the generation process of LMs while balancing control flexibility, control granularity, and generating efficiency remains an open challenge. RMT shows potential in addressing these issues, making it a compelling solution for controllable text generation.

\subsection{RMT v.s NRP}

The most similar to RMT is Non-Residual Prompt (NRP) approach\cite{carlsson-etal-2022-fine}, which also requires pretraining and fine-tuning. Therefore, we make a thorough discussion of differences of RMT w.r.t NRP.

\begin{itemize}

\item \textbf{Principle reasonability.} The principle of RMT appears intuitively more reasonable. RMT utilizes cross-attention to uniformly apply control conditions to each generated token, alleviating the ``distance bias'' problem that arises with increasing distance from the control prompts. By contrast, NRP first learns the fixed position-invariant values and then combines non-residual attention (i.e., causal attention) to address this issue. Intuitively, NRP still uses distance-aware causal attention to encode control conditions, and seems inherently difficult to fully overcome the ``distance bias'' issue. In contrast, cross-attention in RMT enables the final representation of the current generation step to look up the control condition position-independently, which is more intuitively appealing.

\item \textbf{Training efficiency.} RMT holds a clear advantage over NRP in computational source cost. RMT efficiently reuses the output of the base CLM, requiring fewer pretraining training data and expediting the pretraining process. Additionally, RMT is lightweight and is built on the top layer of the base-CLM. Hence, the backpropagation process does not need to propagate into the CLM, which will save a huge training time and GPU memory. Conversely, NRP, despite being initialized from the base CLM, necessitates training a separate base-CLM and involves multi-stage pretraining, leading to higher computational costs. As discussed in Appendix A.1, RMT's pre-training was conducted on a single NVIDIA A6000 GPU, completed in approximately 30 hours. In contrast, according to the report in the paper on NRP, they incurred approximately 400 GPU hours (using DGX-100 machines) for their training process.

\item \textbf{Parameter efficiency.} RMT is more parameter-efficient compared to NRP. NRP requires training additional base-CLM, whereas RMT only necessitates approximately 1/6 of the base-CLM's parameters.

\end{itemize}

\section{Human Evaluation}
\label{human_evaluation}
We conduct the human evaluation for the word inclusion and sentiment control tasks. To ensure the reliability of the evaluation, we invite three highly educated experts as evaluators, and each evaluator has expertise in English. For the evaluation process, we establish a unified criterion using a scale of 10 for each aspect. The scale ranges from 1 to 3 for poor performance, 4 to 7 for satisfactory performance, and 8 to 10 for excellent performance. This criterion allows evaluators to provide consistent and comparable scores for each aspect.

Regarding the CommonGen dataset and C2Gen, we randomly select 25 samples from test dataset for each CTG approach, resulting in a total of 25*7 =175 samples for CommonGen and 50 samples for C2Gen. For the sentiment control experiments, we create 10 positive-steering prompts composed of neutral and positive prompts, as well as 10 negative-steering prompts composed of neutral and negative prompts. Each prompt is evaluated using three CTG approaches, resulting in a total of 60 samples for the final human evaluation. To obtain the final results, we calculate the average score for each metric for every CTG method under comparison, based on the evaluations provided by the experts.

\section{Qualitative Examples}
\label{Qualitative_Examples}

In this section, we list some examples generated by different CTG approaches. Specifically, Table~\ref{app:common_gen_examples} shows the cases on the CommonGen dataset, and Table~\ref{c2gen_examples} presents the cases in the C2Gen dataset. As for attribute control, we show the examples on the adversarial steering setting (i.e., the polarity of the prompt's sentiment is opposite from the target sentiment) as shown in table~\ref{sentiment_example}.  

%% ***********************************************************************
%% ***********************************************************************
% \begin{table}[htbp]
\begin{table*}[htbp]
\centering
\renewcommand{\arraystretch}{2}
% \resizebox{\linewidth}{!}{
\begin{tabular}{lll}
\toprule[1pt]
\multicolumn{3}{l}{\textbf{Target Words}: \color{red}{swim}, \color{blue}{pool}, \color{cyan}{reach}, \color{magenta}{end}}                                        \\ \midrule[0.5pt]
%% ****************************************
POINT~\cite{zhang-etal-2020-pointer} & \makecell[l]{water at the south \textcolor{magenta}{end} of the main \textcolor{red}{swimming} \textcolor{blue}{pool} allows one to \\ \textcolor{cyan}{reach} to the top where they can then swim to the main pool.}
\\ \hdashline 
%% ****************************************
NeuroLogic~\cite{lu-etal-2022-neurologic} & \makecell[l]{The \textcolor{magenta}{end} of the world is coming, and it will be \textcolor{red}{swimming} in blood \\ \textcolor{blue}{pools} of blood, and the blood will be so thick that it will \textcolor{cyan}{reach} the sky}
\\ \hdashline 
%% ****************************************

NRA~\cite{carlsson-etal-2022-fine} & \makecell[l]{The \textcolor{red}{swim} \textcolor{blue}{pool} at the \textcolor{magenta}{end} of the \textcolor{blue}{pool} can be used to \textcolor{cyan}{reach} the other \\ side of the building.} \\ \hdashline 
%% ****************************************
ChatGPT & \makecell[l]{I can't wait to \textcolor{red}{swim} in the \textcolor{blue}{pool} and \textcolor{cyan}{reach} the \textcolor{magenta}{end}.} \\ \hdashline 
%% ****************************************
RMT (ours) & \makecell[l]{The \textcolor{red}{swimming} \textcolor{blue}{pool} has \textcolor{cyan}{reached} its maximum size and is at the \textcolor{magenta}{end} \\ of its life-spans.}                                       
\\ \toprule[1pt]
%% ****************************************
%% ****************************************
\multicolumn{3}{l}{\textbf{Target words}: \color{red}{look}, \color{blue}{heart}, \color{cyan}{writer}, \color{magenta}{leave}, \color{brown}{goer}}
\\ \midrule[0.5pt]
%% ****************************************
%% ****************************************
POINT~\cite{zhang-etal-2020-pointer}  & \makecell[l]{i \textcolor{brown}{goer} in my \textcolor{blue}{heart}, but you will not \textcolor{magenta}{leave} me alone, do you \textcolor{red}{look} \\ like a \textcolor{cyan}{ writer}?.}   
\\ \hdashline 
%% ****************************************
KG-BART~\cite{liu2021kg} & \makecell[l]{a group of people \textcolor{red}{looking} at each other and \textcolor{magenta}{leaving} their \textcolor{blue}{hearts}\\  in the shape of a \textcolor{blue}{heart} with a \textcolor{cyan}{writer}}
\\ \hdashline 

%% ****************************************
NeuroLogic~\cite{lu-etal-2022-neurologic} & \makecell[l]{The first thing I did was \textcolor{red}{look} at the \textcolor{brown}{goers} and the \textcolor{blue}{heart} rate, \\ and I \textcolor{magenta}{left} the room and went to the \textcolor{cyan}{writers} room.}
\\ \hdashline 

%% ****************************************
NRA~\cite{carlsson-etal-2022-fine} & \makecell[l]{The \textcolor{cyan}{writer}-producer would \textcolor{magenta}{leave} the show to \textcolor{red}{look} for a new job \\ as a \textcolor{brown}{goer}.}                                         
\\ \hdashline 
%% ****************************************
ChatGPT & \makecell[l]{The \textcolor{cyan}{writer}'s \textcolor{blue}{heart} sank as they watched the \textcolor{brown}{goer} \textcolor{magenta}{leave} without \\ a second \textcolor{red}{look}.}      
\\ \hdashline 
%% ****************************************
RMT (ours) & \makecell[l]{The \textcolor{cyan}{writer} \textcolor{magenta}{leaves} a \textcolor{blue}{heart}-shaped letter for the \textcolor{brown}{goers} to \textcolor{red}{look} at on \\ the red carpet.}                               
\\ 
%% ****************************************
\bottomrule[1pt]
\end{tabular}
% }
\vspace{0.3cm}
\caption{The examples generated by different CTG approaches on the CommonGen dataset (the without-context setting).}
\label{app:common_gen_examples}
\end{table*}

\begin{table*}[htbp]
\renewcommand{\arraystretch}{2}
% \resizebox{\linewidth}{!}{
\begin{tabular}{ll}
\toprule[1pt]
\multicolumn{2}{c}{\textbf{Target Words \& Context}} 
\\ \midrule[0.5pt]
% \textcolor{red/blue/green/black/white/cyan/magenta/yellow}{text}
%% ********************************************
1 & \begin{tabular}[c]{@{}l@{}}\textbf{Target words:} \textcolor{red}{enjoy}\#\textcolor{red}{friend}\#\textcolor{red}{knit}\\ \textbf{Context:} Friendship is a relationship of mutual affection between people. It is a stronger form of \\ interpersonal bond than an association, and has been studied in various academic fields. Such as \\ communication, sociology, social psychology, anthropology, and philosophy.\end{tabular} \\ \hdashline % \hline

%% ********************************************
2 & \begin{tabular}[c]{@{}l@{}}\textbf{Target Words:} \textcolor{blue}{background}\#\textcolor{blue}{crowd}\#\textcolor{blue}{pose}\\ \textbf{Context:} I love photography. One of my favorite things to do is to take pictures of people modeling \\ throughout the city. The beach is one of my favorite places because of the scenery and the \\ atmosphere there.\end{tabular}  \\ \hdashline  % \hline
%% ********************************************
3 & \begin{tabular}[c]{@{}l@{}}\textbf{Target Words:} \textcolor{brown}{cigarette}\#s\textcolor{brown}{street}\#\textcolor{brown}{walk}\\ \textbf{Context:} James mulled over the fight with his girlfriend. The city noises were all blocked out by \\ him. He dropped his lighter and then picked it up again.\end{tabular}  
\\ \toprule[1pt]
%% ********************************************
%% ********************************************
\multicolumn{2}{c}{\textbf{ChatGPT}}       
\\ \midrule[0.5pt]
%% ********************************************
%% ********************************************
1 & My \textcolor{red}{friend} and I \textcolor{red}{enjoy} spending time together, often \textcolor{red}{knitting} and chatting.\\ \hdashline   %\hline
2 & The \textcolor{blue}{background} of the beach makes for a stunning setting for a \textcolor{blue}{crowd pose}. \\ \hdashline  %\hline
3 & As he lit his \textcolor{brown}{cigarette}, he decided to take a \textcolor{brown}{walk} down the \textcolor{brown}{street}. 
\\ \toprule[1pt]
%% ********************************************
%% ********************************************
\multicolumn{2}{c}{\textbf{RMT}}
\\ \midrule[0.5pt]
%% ********************************************
%% ********************************************
1 & Friendship can be \textcolor{red}{enjoyed} by both men and women and it is best \textcolor{red}{knit} between two close \textcolor{red}{friends}.  \\ \hdashline  %\hline
2 & It is so much fun to \textcolor{blue}{pose} in the \textcolor{blue}{background} with a \textcolor{blue}{crowd} of people.  \\ \hdashline  %\hline
3 & As he \textcolor{brown}{walked} up to the \textcolor{brown}{street}, his \textcolor{brown}{cigarettes} were gone from his mouth.  
\\ \bottomrule[1pt]
\end{tabular}
% }
\vspace{0.2cm}
\caption{The examples generated by different CTG approaches on the C2Gen dataset (the with-context setting).}
\label{c2gen_examples}

\end{table*}

\begin{table*}
\renewcommand{\arraystretch}{2}
\resizebox{\linewidth}{!}{
\begin{tabular}{ll}
\toprule[1pt]
\multicolumn{2}{l}{\textbf{\textcolor{blue}{Negative Prompt}:} \textit{The real problem, though, is that we don} (\textbf{\textcolor{red}{Positive Steering}})}                                   
\\ \midrule[0.5pt]
PPLM~\cite{pplm}                                & \begin{tabular}[c]{@{}l@{}}\textit{The real problem, though, is that we don}'t have a sensible way of using them to share \\ meta-information. The investigation, no evidenced-\end{tabular}
\\ \hdashline
%%% ***********************************
CTRL~\cite{ctrl}                                & \begin{tabular}[c]{@{}l@{}}\textit{The real problem, though, is that we don}'t know how long the product has been in use.\\  It should have been tested for freshness and potency befor\end{tabular} 
\\ \hdashline
%%% ***********************************
GEDI~\cite{gedi}                                & \begin{tabular}[c]{@{}l@{}}\textit{The real problem, though, is that we don}'t teach children all of these tools.  \\ And so what can parents do to help them acquire all this\end{tabular}           
\\ \hdashline
%%% ***********************************
DEXPERT~\cite{liu-etal-2021-dexperts}                             & \begin{tabular}[c]{@{}l@{}}\textit{The real problem, though, is that we don}'t know the full extent of what social causes \\ contribute  to obesity and other related health conditions, says study \end{tabular}                  
\\ \hdashline
%%% ***********************************
DisCup~\cite{zhang-song-2022-discup}                      & \begin{tabular}[c]{@{}l@{}}\textit{The real problem, though, is that we don}'t know what to do with our children but that's  \\ OK too because the world has to grow but 
\end{tabular}                  
\\ \hdashline
%%% ***********************************
RMT (Ours)                       & \begin{tabular}[c]{@{}l@{}}\textit{The real problem, though, is that we don}'t question whether 'good' or 'bad' will be better \\ but keep putting our faith into not\end{tabular}                         
\\ \toprule[1pt]
%%% **************************************************************
%%% **************************************************************
\multicolumn{2}{l}{\textbf{\textcolor{red}{Positive Prompt}:} \textit{A President unafraid to stand} \textbf{(\textcolor{blue}{Negative Steering})}}  
\\ \midrule[0.5pt]
%%% ***********************************
PPLM~\cite{pplm}                                & \begin{tabular}[c]{@{}l@{}}\textit{A President unafraid to stand} out: No matter what the circumstances, Trump will never be\\ the first president to fight one
\end{tabular}                  
\\ \hdashline
%%% ***********************************
CTRL~\cite{ctrl}                                & \begin{tabular}[c]{@{}l@{}}\textit{A President unafraid to stand} up for what he believes in. This one could be added to \\ my library.  Rating: 2.0  I\end{tabular}    
\\ \hdashline
%%% ***********************************
GEDI~\cite{gedi}                                 & \begin{tabular}[c]{@{}l@{}}\textit{A President unafraid to stand} by NRA shameless hypocrites tweeted something totally me \\ aningless that called a black woman a `rapist'. Neither hurtful\end{tabular}   
\\ \hdashline
%%% ***********************************
DEXPERT~\cite{liu-etal-2021-dexperts}                              & \begin{tabular}[c]{@{}l@{}}\textit{A President unafraid to stand} on a bomb, is he?That was bad enough to threaten  \\ the security of every armed\end{tabular}   
\\ \hdashline
%%% ***********************************
DisCup~\cite{zhang-song-2022-discup}                              & \begin{tabular}[c]{@{}l@{}}\textit{A President unafraid to stand} alone? I'm not a good person or good \\ for a President to be? And even if I am\end{tabular}    
\\ \hdashline       
%%% ***********************************
RMT (ours)                          & \begin{tabular}[c]{@{}l@{}}\textit{A President unafraid to stand} up against corruption? Unfortunately, \\ there appears little hope right now of getting any meaningful changes. Since Congress \end{tabular}     \\ \bottomrule[1pt]
\end{tabular}
}

\vspace{0.3cm}
\caption{The generation examples of typical CTG approaches in the sentiment control task. We deliberately select two challenging prompts, and ask the above CTG approaches to conduct the adversarial steering of sentiment polarity.}
\label{sentiment_example}
\end{table*}

\end{document}